\pdfoutput=1 % arXiv: must stay within the first 5 lines so AutoTeX uses pdfLaTeX
\documentclass[letterpaper]{article}
\usepackage[preprint]{aaai2027-arxiv}
\usepackage[hyphens]{url}
\usepackage{graphicx}
\usepackage{natbib}
\usepackage{caption}
\usepackage{algorithm}
\usepackage{algorithmic}

\usepackage{newfloat}
\usepackage{listings}
\DeclareCaptionStyle{ruled}{labelfont=normalfont,labelsep=colon,strut=off}
\floatstyle{ruled}
\newfloat{listing}{tb}{lst}{}
\floatname{listing}{Listing}

\usepackage{booktabs}
\usepackage{tikz}
\usetikzlibrary{arrows.meta, positioning, calc}
\usepackage{amsmath}
\usepackage{amssymb}

\usepackage[hyperfootnotes=false]{hyperref}
\definecolor{qsvlink}{RGB}{0,0,160}
\hypersetup{
    colorlinks=true,
    linkcolor=qsvlink,
    citecolor=qsvlink,
    urlcolor=qsvlink,
    pdftitle={QSV: Quat-Sphere-Vision for Coupled Quaternion Attention on Spherical Lattices},
    pdfauthor={Nicholas Foley, Devin Marinelli, Donny Moore, Diego Enriquez, Amanda Fernandez},
}

\title{QSV: Quat-Sphere-Vision for Coupled Quaternion Attention on Spherical Lattices}

\author {
    Nicholas Foley,
    Devin Marinelli,
    Donny Moore, \\
    Diego Enriquez,
    Amanda Fernandez
}
\affiliations {
    Department of Computer Science, University of Texas at San Antonio\\
    nicholas.foleyblanchette@utsa.edu, devin.marinelli@utsa.edu, donald.moore2@utsa.edu, \\
    diego.enriquez@utsa.edu, amanda.fernandez@utsa.edu
}

\begin{document}

\maketitle

\begin{abstract}
In standard attention, three separately learned projections decide how strongly a token attends to each neighbor ($W_Q$, $W_K$) and how the attended features are transformed before aggregation ($W_V$). We study Quat-Sphere-Vision (QSV),
a sparse spherical vision model that replaces this projection triple with a single learned
unit quaternion per token: the relative quaternion $r_{ij} = q_i^{*} \otimes q_j$
supplies both the attention logit $\operatorname{Re}(r_{ij})$ and a sandwich-product feature
transport $x \mapsto r_{ij} \otimes x \otimes r_{ij}^{*}$, with messages passed
over sparse kNN graphs on concentric Fibonacci spheres. Ablations that change only the targeted component show the two roles to be asymmetric. Removing the transport reduces test accuracy by about four percentage points on CIFAR-10 and CIFAR-100 (single runs per CIFAR-100 variant), while replacing the learned attention weights with uniform averaging leaves it
essentially unchanged. Parameter-matched controls then remove the geometry itself: standard attention on the same graph exceeds QSV (mean $87.3\%$ vs.\ $85.9\%$),
and the same model on a flat 2D lattice reaches $91.1\%$, within $2.1$ points of a ResNet-20 trained under the same pipeline (single run).
In the coupled kernel, nearly all of the learned pairwise computation resides in the transport channel.
\end{abstract}

\section{Introduction}
\label{sec:intro}

The dot product attention operator in the standard Vision Transformer~\cite{dosovitskiy2021vit,vaswani2017attention} parameterizes the interaction between a pair of tokens with three independent linear projections. Two of them, $W_Q$ and $W_K$, produce vectors whose inner product determines an attention weight; the value projection $W_V$ determines what is communicated once the weight is applied. The two functions, how strongly to attend and how to transform the message, are learned by separate parameter sets with no shared structure.

\begin{figure}[t]
\centering
\includegraphics[width=\columnwidth]{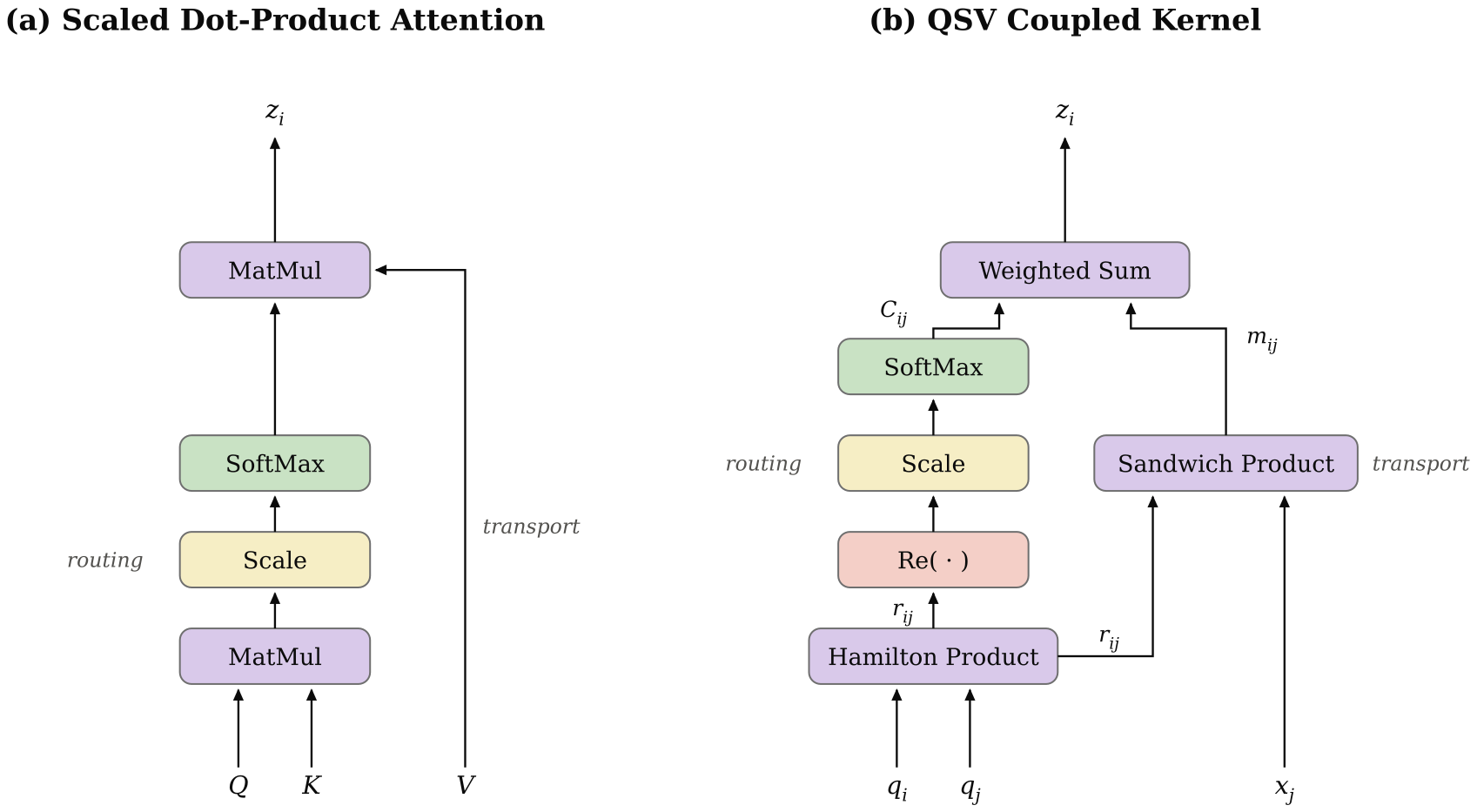}
\caption{(a)~Scaled dot-product attention. (b)~The QSV coupled kernel:
one relative quaternion $r_{ij} = q_i^{*} \otimes q_j$
(Eq.~\eqref{eq:rel-quat}) yields both the attention weight
($\Re(r_{ij})$ scaled by a learned $\beta_h$) and the message transport
$m_{ij} = r_{ij} \otimes \mathbf{x}_j \otimes r_{ij}^{*}$;
the output is $\mathbf{z}_i = \sum_j C_{ij}\, m_{ij}$
(\S\ref{sec:attention}).}
\label{fig:kernel}
\end{figure}

Quat-Sphere-Vision (QSV) replaces the $\{W_Q, W_K, W_V\}$ triple inside the attention kernel with a single unit quaternion $q_i \in \mathbb{S}^3$ attached to each token $i$. For any pair $(i, j)$, the relative quaternion $r_{ij} = q_i^{*} \otimes q_j$ (Eq.~\eqref{eq:rel-quat}) does two tasks. The real part $\Re(r_{ij}) = \cos(\theta_{ij}/2)$ measures alignment between $q_i$ and $q_j$ on $\mathbb{S}^3$ and computes an unnormalized attention score, while the sandwich product $x \mapsto r_{ij} \otimes x \otimes r_{ij}^{*}$, which rotates the imaginary part of $x \in \mathbb{H}$, replaces the value transform. A single learned quaternion per token thus couples ``how much to attend'' and ``what to do with the message'' in one geometric object; Figure~\ref{fig:kernel} contrasts the two kernels.

Tokens are placed at vertices of a Fibonacci lattice on $L$ concentric spheres in $\mathbb{R}^3$ rather than a 2D image grid. The ViT linear patch projection is replaced by a Lambert projection from each first shell vertex back to image coordinates, where a small $5\times5$ kernel of pixels is sampled bilinearly. Connectivity between vertices is sparse; each target aggregates only from its $k$ nearest neighbors, yielding an edge ratio well below the $O(V^2)$ baseline. The conventional per token linear layers are replaced by Hamilton structured linears and a quaternion FFN block, and the unit quaternion parameters are trained with Riemannian Adam on $\mathbb{S}^3$ (\S\ref{sec:training}).

The two kernel roles are far from equally important. The transport is essential: replacing it with an identity message reduces test accuracy by $4.3$ and $4.1$ points on CIFAR-10 and CIFAR-100 respectively. The routing weight derived from the same quaternion is not, since replacing the learned softmax over $\Re(r_{ij})$ with uniform aggregation changes accuracy by at most $0.4$ points on both datasets. Routing is dispensable only given transport, however: disabling both roles at once, or freezing the quaternion field at random values, collapses accuracy to about $65\%$, so the learned field must feed at least one role, and transport is the stronger channel. Two matched controls then isolate what remains. A parameter- and graph-matched dot-product-attention baseline (\S\ref{sec:exp-baseline}) performs at least as well ($87.3\%$ vs.\ $85.9\%$ mean), and replacing the spherical layout itself with a flat 2D lattice at matched parameters reaches $91.1\%$, within about two points of a ResNet-20 trained under the same pipeline. The transport ablation that costs the quaternion model $4.3$ points costs matched attention $1.0$ point on either layout (\S\ref{sec:exp-grid}).

We make three contributions. (i) Through matched ablations, a $2{\times}2$ factorization of the kernel, and two matched controls (standard attention on the same graph; the same pipeline on a flat lattice), we show that, on CIFAR-10/100 at this model scale, the learned transport channel is what matters, while the routing weight (given transport), the quaternion form of the kernel, and the spherical layout itself are each replaceable, the last at a $3.8$-point gain in the matched attention variant. (ii) As the testbed for this study, we construct a sparse multi-shell spherical image model with learned quaternion-valued vertex frames and a message-passing kernel in which one relative rotation provides both a routing logit and a feature transport. (iii) We provide a diagnostic analysis indicating that the bottleneck is generalization on fine local texture, much of it attributable to the spherical layout itself.
\section{Related Work}
\label{sec:related}

\subsection{Quaternion attention models} Quaternion valued networks~\cite{parcollet2018qcnn,parcollet2020survey} have been extended to attention~\cite{tay2019quaternion} and to graph message passing~\cite{nguyen2021qgnn}. These architectures retain the $\{Q, K, V\}$ separation and operate on quaternion valued features without coupling the routing weight to the value transform. \citet{yamauchi2026quaternion} report that the four component-wise scores of quaternion self-attention effectively span the same interaction subspace as a single shared real score; our uniform-routing ablation pushes a step further in this direction: even the single quaternion-derived score can be replaced by uniform aggregation at no cost, given the transport channel (\S\ref{sec:exp-ablation}). The nearest ``one quaternion supplies both a transform and a score'' precedent is QuatE~\cite{zhang2019quate}, where a per-relation quaternion acts on the entity embedding by a one-sided Hamilton product and the score is an inner product; QSV differs in using a two-sided sandwich transport and a per-vertex rather than per-relation quaternion inside a multi-layer kernel.

\subsection{Equivariant attention and capsule routing} Equivariant attention~\cite{fuchs2020se3,brehmer2023gatr,liao2023equiformer} preserves group symmetry, but produces attention weights and equivariant values from separate parameterizations: an invariant weight deciding \emph{how much} to aggregate and an equivariant kernel deciding \emph{how} to transform. Our ablations probe this division of labor in a model that couples the two. The sandwich product itself is a standard equivariant transport for quaternion features~\cite{shen2020rotationequiv}, where the rotation is data-derived to guarantee equivariance; in QSV it is a free learned parameter and no equivariance is claimed. QSV's routing structure is most similar to capsule networks~\cite{sabour2017capsules} and Quaternion Equivariant Capsule Networks~\cite{zhao2020qecnet}, which route between quaternion valued local frames on point clouds via a Weiszfeld iteration. Several studies find capsule routing itself dispensable: uniform or fixed coupling matches learned agreement routing~\cite{paik2019capsules,gu2020capsules,byerly2021norouting}. Our routing result extends this negative finding to a non-iterative, quaternion-alignment routing weight.

\subsection{Rotation based positional encoding} Methods like RoPE and its variants~\cite{su2024rope,heo2024ropevit} rotate $Q$ and $K$ but leave $V$ untouched, and do not derive a routing weight from the rotation itself. Concurrent GeoPE~\cite{yao2025geope} applies the sandwich product in attention on the $Q$/$K$ path as a positional bias, with rotations built from coordinates; no learned rotation or value-path transport is reported. RiemannFormer~\cite{ji2025riemannformer} likewise transports only $Q$ and $K$ (QSV's sandwich is, in effect, the $\mathrm{SO}(3)$ special case of its $\mathrm{SO}(4)$ factorization) while values are aggregated as a standard linear sum. This is the opposite design choice, and our ablations test it directly: the geometric structure helps in the value path and has no measurable effect in the score path. Concurrent RoVE~\cite{garciacastellanos2026rove} points in a similar direction, extending rotary embeddings to the value pathway in language modeling; and concurrent CliffordNet~\cite{ji2026cliffordnet} argues a geometric-product kernel is itself the contribution on CIFAR without, to our knowledge, a parameter-matched attention control; our study supplies that control in our setting and finds the matched kernel performs at least as well. QSV instead derives the routing logit $\Re(r)$ and the value transform from the same learned per-token $r$, eliminating the $\{Q, K, V\}$ triple inside the kernel.

\subsection{Spherical and gauge-equivariant networks} QSV's token layout connects to spherical CNNs~\cite{cohen2018sphericalcnn,cohen2019gauge} and to vision transformers on spherical lattices~\cite{carlsson2024healswin}. Among these, the gauge-equivariant family~\cite{cohen2019gauge,dehaan2021gem}, which transports features between per-vertex frames using a \emph{fixed} geometric connection derived from the mesh; the transport in QSV is instead a \emph{learned} quaternion connection between per-vertex frames, and we make no equivariance claim for it. 

\subsection{Token mixers and the role of attention weights} Our matched-attention result instantiates, in a geometric setting, the observation that the surrounding architecture often matters more than the specific token mixer: pooling, MLPs, or convolutions can replace attention in the standard ViT architecture with little loss~\cite{yu2022metaformer,tolstikhin2021mlpmixer}, and PointMixer extends this to irregular point neighborhoods~\cite{choe2022pointmixer}. A parallel line finds the attention \emph{weights} themselves dispensable or near-uniform in transformers~\cite{tay2021synthesizer,dong2021rank}, graph attention~\cite{brody2022gatv2}, and ViTs~\cite{raghu2021vision,hyeonwoo2023scratching}; in language, frozen random query--key routing stays competitive while the value/MLP path remains critical~\cite{dong2025mixit}. Attention-entropy work treats sharpness extremes as pathologies to fix~\cite{zhai2023stabilizing}; in our setting routing sharpness is inconsequential. Relative to this line, our decomposition is finer grained: rather than swapping the whole mixer, we decompose one geometric kernel into its routing weight and its feature transport, and show that the routing half is dispensable given the transport channel (consistent with the work above; \S\ref{sec:exp-ablation}) while the transport half is not. Its specific algebraic form, however, is interchangeable with a matched attention kernel.
\section{Methods}
\label{sec:models}

A unit quaternion $q \in \mathbb{S}^3$ acts on $v \in \mathbb{R}^3$ by the sandwich product $v \mapsto q \otimes v \otimes q^{*}$. The Lie algebra $\mathfrak{su}(2) \cong \mathbb{R}^3$ has exponential map $\exp(\omega) = (\cos\|\omega\|,\, \sin\|\omega\|\,\omega/\|\omega\|)$, used for parameter updates and for data dependent perturbations of the base quaternions.

\begin{figure}[t]
\centering
\includegraphics{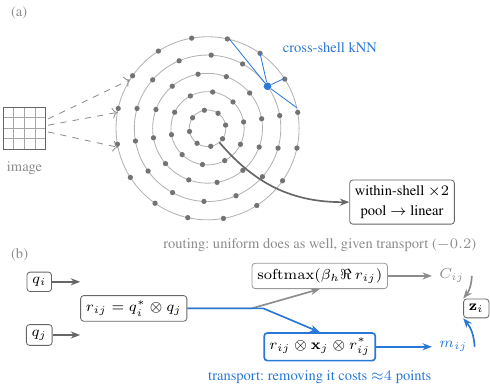}% pre-rendered TikZ
\caption{QSV in one view. (a)~Images are sampled onto the outer shell by the
Lambert patch embedding, propagated inward over sparse cross-shell kNN edges
through five concentric Fibonacci shells, aggregated in two within-shell rounds
on the last shell, and pooled. (b)~The coupled kernel: one relative quaternion
$r_{ij}$ (Eq.~\ref{eq:rel-quat}) supplies both the routing weight and the
feature transport of Eq.~\ref{eq:attn-weight}; the labeled costs are the
ablation results of \S\ref{sec:exp-ablation}.}
\label{fig:arch}
\end{figure}

\subsection{Token layout and geodesic patch embedding}
\label{sec:layout}

Tokens are placed at vertices of $L = 5$ nested spheres (Figure~\ref{fig:arch}) of radii decreasing linearly from $1.0$ to $0.4$, with $V_l = \mathrm{round}\!\big(V_1 (1 + \tfrac{1}{2}(l-1))\big)$ vertices on shell $l \in \{1, \dots, L\}$ ($V_1 = 128$), giving $V = [128, 192, 256, 320, 384]$; initial positions follow a Fibonacci lattice~\cite{saff1997distributing} and are stored as learnable parameters in $\mathbb{R}^3$.

The first shell lives in $\mathbb{R}^3$, so the standard ViT linear patch projection~\cite{dosovitskiy2021vit} cannot be applied.
Each first shell vertex $\mathbf{p}_i$ instead drives a Lambert projection back to image coordinates,
\(u_i = \arctan2(z_i, x_i)/\pi,\; v_i = y_i\) (after normalizing $\mathbf{p}_i$).
A $5\times5$ kernel of offsets at scale $2/\sqrt{V}$ around $(u_i, v_i)$ is sampled by bilinear interpolation with border padding; the $5{\times}5{\times}C$ samples are flattened and projected linearly to dimension $d$, then layer normed.
This projection is differentiable in $\mathbf{p}_i$, allowing gradients to flow through into first shell positions.

\subsection{Quaternion attention}
\label{sec:attention}

Every shell consists of $V_l$ positions and a tensor of base quaternions $q_i^{(l, h)} \in \mathbb{S}^3$ for vertex $i$ and head $h \in \{1, \dots, H\}$. Features at vertex $i$ are reshaped from $\mathbb{R}^d$ into a quaternion valued tensor $\mathbf{x}_i \in \mathbb{H}^{H \times Q}$ with $d = HQ \cdot 4$.

For target vertex $i$ and source vertex $j \in \mathrm{NN}_k(i)$ (\S\ref{sec:connectivity}), the kernel is built around the relative quaternion
\begin{equation}
\label{eq:rel-quat}
r_{ij} \;=\; q_i^{*} \otimes q_j,
\end{equation}
where $q^{*}$ denotes the quaternion conjugate. The per head attention weight $C_{ij}$, message $m_{ij}$, and aggregated output $\mathbf{z}_i$ are all derived from this single object:
\begin{equation}
\label{eq:attn-weight}
\begin{aligned}
C_{ij} &= \mathrm{softmax}_{j}\!\left(\beta_h \, \Re(r_{ij})\right), \\
m_{ij} &= r_{ij} \otimes \mathbf{x}_j \otimes r_{ij}^{*}, \\
\mathbf{z}_i &= \textstyle\sum_{j} C_{ij}\, m_{ij}.
\end{aligned}
\end{equation}
The score $\Re(r_{ij}) = \cos(\theta_{ij}/2) \in [-1, 1]$ is a bounded alignment score; the temperature $\beta_h = \mathrm{softplus}(\hat{\beta}_h) + 0.1$ is learned per head per shell. The transform $m_{ij}$ is the conjugation by $r_{ij}$ acting on the imaginary parts of each channel quaternion of $\mathbf{x}_j$. The same $r_{ij}$ appears in both $C_{ij}$ and $m_{ij}$, so the architecture couples routing and transport through a single geometric object rather than the standard $\{Q, K, V\}$ projection triple.

Post aggregation update applies a Hamilton structured linear followed by the quaternion FFN block introduced below, denoted $\mathrm{Hamilton}_{\mathrm{post}}$, and adds a residual. Writing $\mathbf{u}_i = \mathrm{Hamilton}_{\mathrm{post}}(\mathbf{z}_i)$, the residual is added directly when the shell sizes match, and uses linear interpolation along the vertex axis with a learned gate when they differ:
\begin{equation}
\label{eq:residual}
\mathbf{h}_i^{(l)} =
\begin{cases}
\mathbf{u}_i + \mathbf{x}_i^{(l-1)}, & V_l = V_{l-1},\\[4pt]
\mathbf{u}_i + \sigma(\hat{g}_l)\,\mathrm{interp}(\mathbf{x}_i^{(l-1)}), & V_l \neq V_{l-1}.
\end{cases}
\end{equation}

The full forward pass is: geodesic patch embedding onto the first shell; one
within-shell propagation step on that shell; four cross-shell steps
$\mathcal{S}^{(l-1)} \to \mathcal{S}^{(l)}$, each an application of
Eq.~\eqref{eq:attn-weight} followed by the Hamilton block and residual of
Eq.~\eqref{eq:residual}; two further within-shell aggregation rounds on the last
shell over a geodesic $k{=}16$ self-graph, with their own learned temperatures;
and the quaternion-native pooling of \S\ref{sec:training}.

\subsubsection{Lie algebra perturbation} A static $q_i$ would make the kernel input independent, so a data-dependent perturbation in $\mathfrak{su}(2)$ is added. A per shell, per head MLP $\phi_h\colon \mathbb{R}^d \to \mathbb{R}^3$ produces a tangent vector that perturbs the base quaternion,
\begin{align}
\omega_i^{h} &= \varepsilon_l \tanh(\phi_h(\mathbf{x}_i)), \label{eq:perturb-omega}\\
q_i^{\mathrm{eff}} &= \mathrm{normalize}(q_i^{\mathrm{base}} \otimes \exp(\omega_i^{h})), \label{eq:perturb}
\end{align}
with $\varepsilon_l \in (0.01, \pi/4)$ a learned per shell upper bound, and Eq.~\eqref{eq:rel-quat} uses $q^{\mathrm{eff}}$. The MLP's last layer is zero initialized, making the perturbation identity at the start of training.

\subsubsection{Hamilton structured linears}\label{sec:hamilton} A standard linear layer mixes the four components of each quaternion freely, without preserving the quaternion structure. Hamilton structured linears instead use weights that are themselves quaternions~\cite{parcollet2018qcnn}, $W \in \mathbb{H}^{H \times Q_{\mathrm{out}} \times Q_{\mathrm{in}}}$ with 
\begin{equation}
\label{eq:hamilton}
y_{ho} = \sum_i W_{hoi} \otimes x_{hi},
\end{equation}
implemented as a single dense matmul after expanding each quaternion weight into its $4 \times 4$ real block representation. This is equivalent to the $n=4$ special case of parameterized hypercomplex multiplication~\cite{zhang2021phm}, but with the Hamilton multiplication table fixed by design rather than learned. The ViT FFN $\mathrm{Linear}(d, 4d) \to \mathrm{GELU} \to \mathrm{Linear}(4d, d)$ is replaced by $\mathrm{Hamilton}(Q, 4Q) \to \mathrm{GELU} \to \mathrm{Hamilton}(4Q, Q)$, with the down projection initialized small so the residual path dominates early in training.

\subsection{Sparse kNN connectivity}
\label{sec:connectivity}

The set $\mathrm{NN}_k(i)$ is built once per forward pass from the current vertex positions~\cite{zhao2021pointtransformer}. Cross shell propagation $\mathcal{S}^{(l-1)} \to \mathcal{S}^{(l)}$ uses Euclidean kNN and the final within shell rounds angular kNN; the two select identical neighbors here: every vertex lies exactly on its shell's radius, and between fixed radii $r_1, r_2$ the squared Euclidean distance $r_1^2 + r_2^2 - 2 r_1 r_2 \cos\theta$ is strictly increasing in the angular separation, so both metrics induce the same ranking (confirmed on trained checkpoints). Each shell uses a neighbor count
\begin{equation}
\label{eq:knn-budget}
k_l = \max\!\big(2,\; 2 \lfloor \log_2 V_l \rfloor \big),
\end{equation}
giving $K = [14, 14, 16, 16, 16]$ for the shell sizes above. This yields an edge ratio $|\mathrm{edges}_{\mathrm{kNN}}| / |\mathrm{edges}_{\mathrm{full}}| \approx 0.05$, where the denominator counts each shell's dense within-shell edges.

\subsection{Optimization on $\mathbb{S}^3$ and the shells}
\label{sec:training}

Quaternion parameters are updated by Riemannian Adam~\cite{becigneul2019rieadam,bonnabel2013riemannian}; everything else (positions, MLP weights, $\hat{\varepsilon}$, $\hat{\beta}$, classifier) is updated by AdamW~\cite{loshchilov2019adamw} under a one-cycle, cosine-shaped learning-rate schedule.

\subsubsection{Riemannian Adam} For a quaternion $q$ with Euclidean gradient $g$, the gradient is projected onto the tangent space at $q$ and pulled back to $\mathfrak{su}(2)$ at the identity by the differential of left multiplication by $q$, then a standard Adam step is taken in the Lie algebra and mapped back to the sphere by exp map retraction:
\begin{align}
\omega &= \mathrm{Im}(q^{*} \otimes (g - \langle g, q\rangle q)), \label{eq:rieadam-pullback}\\
\Delta\omega &= -\eta\, \hat{m}/(\sqrt{\hat{v}} + \epsilon), \\
q &\leftarrow \mathrm{normalize}(q \otimes \exp(\Delta\omega)). \label{eq:rieadam}
\end{align}
First and second moments (yielding the bias corrected $\hat{m}$ and $\hat{v}$) are stored in the Lie algebra at the identity, so no parallel transport is required between iterations. The quaternion learning rate is held constant; the moments already produce a natural step size decay, and in preliminary runs annealing the quaternion learning rate froze the field early (not shown).

\subsubsection{Sphere Coulomb tangent repulsion} Without an explicit spreading force, position updates concentrate vertices in low loss regions and deplete coverage near the poles of the sphere. Every $10$ batches a Riesz energy step is applied on each shell: tangential force
\begin{equation}
\label{eq:coulomb}
\mathbf{F}_i = \sum_{j} \frac{1}{\vartheta_{ij}^2 + \epsilon_{\mathrm{r}}}\, \mathbf{t}_{ij}
\end{equation}
from the $k$ nearest neighbors, where $\vartheta_{ij}$ is the angular separation
between vertices $i$ and $j$, $\mathbf{t}_{ij}$ is the unit tangent at
$\mathbf{p}_i$ pointing away from $\mathbf{p}_j$, and $\epsilon_{\mathrm{r}}$ a
small constant. The step is rescaled so that the median angular displacement
equals a decaying target, then each vertex is renormalized to its assigned radius.

\subsubsection{Regularizers and pooling} Three auxiliary losses are added: an antipodal aware quaternion smoothness term $2\xi(1 - |\langle q_t, q_{t-1}\rangle|)$ between consecutive epochs, a per shell head diversity loss $\sum_{h \neq h'} \langle q_i^{(l, h)}, q_i^{(l, h')}\rangle^2$, and an L2 penalty on $\hat{\beta}$. After the last shell, features are pooled by power iterating a mean direction $\bar{q} \in \mathbb{S}^3$ over the last shell base quaternions, rotating $\hat{\mathbf{z}}$ by $\bar{q}$ to produce a 3D direction $\mathbf{d}$, and applying softmax attention over vertices using $\sum_q \mathrm{Im}(\mathbf{x}_i^q) \cdot \mathbf{d}$ as the score; a single linear classifier produces the class logits. Default configuration values are listed in the technical supplement.

\subsection{Datasets}
\label{sec:datasets}

Experiments use \textbf{CIFAR-10} and \textbf{CIFAR-100}~\cite{krizhevsky2009cifar}: each $60{,}000$ color images of size $32\times32$ across $10$ and $100$ classes respectively ($45{,}000$ train, $5{,}000$ val, $10{,}000$ test), with per-channel normalization. CIFAR-100 uses the identical pipeline and configuration apart from the class count and a repulsion neighbor-count adjustment ($8 \to 16$; caption of Table~\ref{tab:ablation}). Throughout the paper, test accuracy is measured at the checkpoint with the best validation accuracy; the test set is never used for model selection.

Training augmentation (padded random crop, horizontal flip, TrivialAugmentWide~\cite{muller2021trivialaugment}) and all remaining hyperparameters are listed in the technical supplement.

\section{Experiments}
\label{sec:exp}

We evaluate the trainability and accuracy of the quaternion kernel, ablate its
two roles to identify which one the model relies on, and study the role of the
Riemannian optimization stack. All experiments use the
default configuration, listed in the technical supplement, on CIFAR-10 unless
stated otherwise.

\subsection{Trainability and accuracy}
\label{sec:exp-main}

Whether the constrained kernel trains stably is not obvious in advance. In practice it does: epoch-level loss decreases near-monotonically and validation accuracy tracks the augmented training accuracy in the default-configuration runs. The model reaches up to $86.9\%$
test accuracy ($87.0\%$ best validation). The result is also stable across runs, with test accuracy staying within $85.7$--$86.9\%$ across a broad sweep
of learning rate, quaternion step size, and repulsion settings under annealed
schedules (a constant learning rate sits about a point lower) and seed variation of up to about one point across the variants of Table~\ref{tab:ablation}.

We do not present these numbers as competitive: Table~\ref{tab:context} provides context and \S\ref{sec:exp-error} locates the remaining gap.

\begin{table}[t]
\centering
\setlength{\tabcolsep}{3.5pt}
\begin{tabular}{lcc}
\toprule
Model & Params & Test acc.\ (\%) \\
\midrule
ResNet-20~\cite{he2016resnet} & 0.27M & 91.3 \\
ResNet-56~\cite{he2016resnet} & 0.85M & 93.0 \\
ViT-Lite-7/4~\cite{hassani2021cct} & 3.7M & 93.6 \\
CCT-7/3$\times$1~\cite{hassani2021cct} & 3.8M & 96.5 \\
\midrule
Matched attention (\S\ref{sec:exp-baseline}) & 0.57M & 87.3 \\
Flat-lattice control (\S\ref{sec:exp-grid}) & 0.57M & 91.1 \\
ResNet-20, our pipeline (\S\ref{sec:exp-grid}) & 0.27M & 93.2 \\
QSV (ours) & 0.64M & 85.9 \\
\bottomrule
\end{tabular}
\caption{Context: small models trained on CIFAR-10. Baseline numbers
are as reported in the cited papers and use different training pipelines and
augmentation; they are included for scale, not as a controlled comparison. Our attention entries are seed means (QSV best run across configurations: $86.9$); the
same-pipeline ResNet-20 is a single run. The two spherical models sit several
points below convolutional baselines; the flat-lattice control closes most of
that gap (\S\ref{sec:exp-grid}).}
\label{tab:context}
\end{table}

\subsection{Which part of the kernel matters}
\label{sec:exp-ablation}

The relative quaternion $r_{ij}$ plays two roles in Eq.~\eqref{eq:attn-weight}:
its real part sets a routing weight and its sandwich action transports features.
To separate them we ablate each role under a matched configuration and seed,
changing only the targeted component. Table~\ref{tab:ablation} reports the result.

\begin{table}[t]
\centering
\setlength{\tabcolsep}{4.5pt}
\begin{tabular}{lcc}
\toprule
Variant & CIFAR-10 $\Delta$ & CIFAR-100 $\Delta$ \\
\midrule
Full kernel (baseline) & $85.9$ (---) & $60.1$ (---) \\
Transport $\to$ identity & $81.7$ ($-4.3$) & $56.0$ ($-4.1$) \\
Routing $\to$ uniform $1/k$ & $85.7$ ($-0.2$) & $59.7$ ($-0.4$) \\
Routing on $|\Re(r)|$ & $86.3$ ($+0.3$) & --- \\
Geometric routing & $86.0$ ($+0.0$) & --- \\
\midrule
Both roles disabled & $64.1$ ($-21.8$) & --- \\
Frozen random field & $66.5$ ($-19.4$) & --- \\
Attention $W_V \to$ identity & $86.3$ ($-1.0$)$^{\dagger}$ & $62.6$ ($-0.8$)$^{\dagger}$ \\
\bottomrule
\end{tabular}
\caption{Kernel ablations under the default configuration on CIFAR-10 and
CIFAR-100. CIFAR-10 entries are mean test accuracy over $2$--$3$ seeds; CIFAR-100
entries are single runs. The change from the full kernel is in parentheses; only
the targeted component changes per row. CIFAR-100 runs differ only in class count and repulsion neighbors ($8 \to 16$; an untuned legacy heuristic, applied to every CIFAR-100 run).
Deltas are from unrounded accuracies; per-seed values are listed in the
appendix. The joint-disable and frozen-field rows are single runs at seed
$42$.
$^{\dagger}$Delta from the matched attention baseline ($87.3$ CIFAR-10 mean;
$63.4$ CIFAR-100 single run; \S\ref{sec:exp-baseline}), not from QSV.}
\label{tab:ablation}
\end{table}

The two roles differ by more than an order of magnitude in their effect on
accuracy, and the asymmetry replicates on a second dataset (one run per
CIFAR-100 variant). Replacing the sandwich transport with an
identity message ($m_{ij} = \mathbf{x}_j$, keeping $C_{ij}$; the rotation acts
on the imaginary components, three of every four feature dimensions) reduces test
accuracy by four points on both CIFAR-10 ($-4.3$) and CIFAR-100 ($-4.1$),
and also lowers the augmented training accuracy (from ${\approx}81\%$ to
${\approx}74\%$), indicating the transport contributes to fitting the (augmented) training data, not only to generalization. Across three seeds the transport-off accuracy is
tight ($81.3$--$82.1$ on CIFAR-10), so the effect is well outside seed noise.
Replacing the learned softmax over $\Re(r_{ij})$ with uniform $1/k$ aggregation,
by contrast, changes accuracy by a few tenths of a point on both datasets
($-0.2$, $-0.4$). Two routing variants behave the same, both within the seed
band: routing on $|\Re(r_{ij})|$, which folds the double cover so antipodal
quaternions score identically, and geometric routing, where the logit is instead
the negative squared distance between the target's direction and the neighbour
direction rotated by the target's quaternion (the original geometric formulation
of the score). The asymmetry is not additive, however. Disabling both roles at
once (uniform routing \emph{and} identity transport) collapses test accuracy to
$64.1$ (about $17$ points below the additive prediction), and freezing the
entire quaternion field at random unit values collapses similarly ($66.5$; both
single runs, Table~\ref{tab:ablation}). With transport removed, the learned
routing weights carry roughly $17$ points; with transport present they carry
almost nothing. In the full model, routing is therefore redundant given
transport rather than useless (consistent with the routing-dispensability
literature of \S\ref{sec:related}); and a frozen rotation field, random or
identity, is no substitute for a learned one. The joint cell also qualifies the pooling-mixer results of \S\ref{sec:related}: uniform aggregation plus per-token blocks is structurally a pooling mixer, yet here it collapses, plausibly because the narrow Hamilton-constrained blocks lack the width and value mixing that make pooling viable in MetaFormer-style architectures.

\subsection{Routing diagnostics}
\label{sec:exp-routing}

\begin{figure}[t]
\centering
\includegraphics[width=0.85\columnwidth]{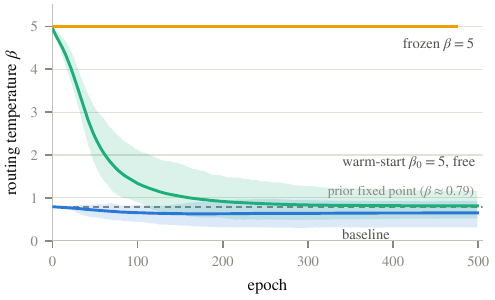}
\caption{Routing-temperature probe (CIFAR-10, seed $42$; one run per
intervention). Lines: means over the logged cross-shell temperatures (four of
$16$ heads per shell); bands: min--max. Warm-started at $\beta_0 = 5$, temperatures decay back to the prior's fixed point ($\beta \approx 0.79$, dashed) at baseline accuracy; freezing $\beta = 5$ throughout costs $-0.4$, within seed noise (\S\ref{sec:exp-routing}).}
\label{fig:beta}
\end{figure}

The ablation is consistent with direct measurement of the routing distributions.
On held-out batches the per-shell attention weights remain close to uniform
throughout training: the normalized entropy stays at about $0.99$ of its maximum
and the peak weight is only $1.2$--$1.4\times$ the uniform value $1/k$, whereas a
sharply selective kernel would peak many times higher. This near-uniformity is
not forced by the parameterization. On a converged checkpoint the
quaternion field is far from collapsed: the within-neighborhood orientation spread
grows from $\approx 0.9$ to $\approx 1.8$ radians across shells, so the routing
scores $\Re(r_{ij})$ carry real variation across a target's neighbors
(per-neighborhood standard deviation $0.24$--$0.46$). The temperature $\beta_h$ is free to grow, subject only to a weak $\ell_2$ prior on its logits (appendix) whose fixed point is $\beta_h \approx 0.79$. A task gradient favoring sharp routing
could easily overcome this weak pull, yet the learned temperatures sit at or
below the prior's fixed point (per-shell head means $\approx 0.50$--$0.76$) and
decrease over training. Two interventions, one run each, probe this directly
(Figure~\ref{fig:beta}). Warm-started at $\beta_h = 5$ (sharp routing made
available), the free cross-shell temperatures return to the regime of the
priors' fixed point (final range $0.52$--$1.16$) and the run finishes at
baseline accuracy ($86.1$ vs.\ $85.9$). However, since the priors alone
would also pull $\beta_h$ back, this shows only that, in this run, no task gradient acts to maintain sharp routing. Freezing $\beta_h = 5$ throughout (sharp routing imposed, free of
that confound; within-shell rounds keep learned temperatures) changes test
accuracy by $-0.4$, within seed noise. Uniform, learned-soft, and forced-sharp
cross-shell aggregation all reach the same accuracy.

\subsection{A matched attention baseline}
\label{sec:exp-baseline}

The ablations above isolate which \emph{part} of the quaternion kernel the model
uses; they do not ask whether the quaternion parameterization is necessary at
all. To test this we replace the entire kernel with standard scaled dot-product
attention (real-valued $\{W_Q, W_K, W_V, W_O\}$ projections and a feed-forward
block per shell) while holding the rest of the model fixed: the same spherical Lambert
tokenizer, multi-shell vertex layout, and sparse cross-shell kNN connectivity
(identical $V$, $K$, radii, and Fibonacci initialization; \S\ref{sec:layout}). The
attention width is set so the baseline ($0.57$M parameters, full model) is
comparable to QSV ($0.64$M), and it is trained with a OneCycle schedule sharing
QSV's peak learning rate, warmup fraction ($10\%$), and epoch horizon; the
baseline runs the full horizon without early stopping, and the reported
checkpoint is selected by validation accuracy, the same selection criterion as
QSV. We train three seeds.

Residual asymmetries favor QSV rather than the baseline: learnable vertex
positions (the baseline's graph is frozen at initialization), two extra
within-shell aggregation rounds with learned pooling versus mean pooling,
dropout and auxiliary regularizers the baseline lacks, and a learning rate
inherited from QSV rather than tuned. The baseline nevertheless matches or
exceeds QSV, so these omissions make the comparison conservative.

The matched attention baseline reaches $87.3\%$ test accuracy (mean over three
seeds, range $87.1$--$87.5$), matching or slightly exceeding both QSV's best run across configurations ($86.9\%$) and the QSV default-configuration mean ($85.9\%$). On CIFAR-100 the same baseline reaches
$63.4\%$ (single run), the reference for the value-path ablation of
Table~\ref{tab:ablation}. Together with the ablation, this parity indicates the quaternion form of the kernel is not the source of the model's performance on this benchmark: transport over the graph matters (an identity message
costs four points, \S\ref{sec:exp-ablation}), while the form of that
transport (quaternion sandwich or learned linear projection) does not. The
quaternion kernel is one workable choice among several, mirroring the
observation that the token mixer often matters less than the surrounding
architecture~\cite{yu2022metaformer} (\S\ref{sec:related}). That leaves the spherical construction itself (\S\ref{sec:exp-grid}).

The two parameterizations allocate parameters differently. At $d=256$ the attention projections cost $1.32$M parameters where QSV's kernel costs about
$87$K ($15\times$ less), letting QSV hold a $256$-dimensional representation
in a $0.64$M budget where standard attention must narrow to $d=96$ (a
same-width $d{=}256$ baseline, $3.97$M parameters, reached only $68.6\%$ in a
single untuned run). On this task that efficiency yields representation width rather than accuracy, so we report it as a property of the construction. Wall-clock cost favors the baseline: on the same RTX A6000 with the identical data
pipeline, the baseline trains at ${\approx}1550$ images/s uncompiled and
${\approx}5350$ compiled, versus ${\approx}470$ for compiled QSV. Beyond being interchangeable, the quaternion kernel is also at least $3.3\times$ (about $11\times$ compiled) more expensive to train.

\subsection{A flat-lattice control}
\label{sec:exp-grid}

To test the spherical construction directly we keep the attention baseline
unchanged (same stage sizes $V$, neighbor counts $K$, $5{\times}5$ sampling
kernel, $0.57$M parameters, schedule, and selection rule) and remove
only the geometry: vertices sit on flat 2D lattices, sampling happens directly
at the lattice coordinates instead of through the Lambert projection, and the
kNN graphs are built in the plane. Across three seeds this control reaches
$\mathbf{91.1\%}$ test accuracy (range $90.9$--$91.3$): $3.8$ points above the
spherical version of the same model ($87.3$, seed ranges non-overlapping) and
within $2.1$ points of a ResNet-20 trained under the same pipeline
($93.2$, single run; Table~\ref{tab:context}). On this task the spherical layout is therefore an active liability, costing $3.8$ points at matched parameters. Combining all controls on CIFAR-10, neither the quaternion kernel, nor the learned routing, nor the spherical layout survives as the source of performance.

A value-path ablation on both attention controls isolates the kernel factor.
Setting the value projection to identity while keeping the attention weights
(the analogue of the identity-message cell of \S\ref{sec:exp-ablation}) costs
the spherical baseline $1.0$ point ($86.3$ mean over three seeds, range
$86.2$--$86.5$; Table~\ref{tab:ablation}) and the flat control $1.0$ point
($90.1$ mean over three seeds, range $90.0$--$90.2$). Standard attention thus loses one point on
either layout, against
$4.3$ for the coupled kernel on the identical spherical graph, a comparison in which the kernel is the principal difference (residual asymmetries of \S\ref{sec:exp-baseline} notwithstanding). The coupled
quaternion kernel concentrates its function in the transport channel, whereas
standard attention distributes it across its remaining learned paths (the
output projection and feed-forward block). The strong transport dependence of
\S\ref{sec:exp-ablation} is thus a property of the coupled kernel rather than of sparse-graph attention in general.

\subsection{Optimization and geometry}
\label{sec:exp-opt}

The Riemannian training stack matters for stability in three ways. Updating the
quaternions with AdamW plus hard renormalization to $\mathbb{S}^3$ trains, but
with noisier loss curves than storing the Adam moments in $\mathfrak{su}(2)$
(Eqs.~\eqref{eq:rieadam-pullback}--\eqref{eq:rieadam}); bounding the
data-dependent perturbation by a learned $\varepsilon_l \in (0.01, \pi/4)$ and
zero-initializing the perturbation MLP's last layer was important for
stability. The sphere Coulomb repulsion, active and decayed over roughly the
first $40\%$ of training, keeps vertex coverage uniform (without it vertices
drift toward concentrated regions within a dozen epochs), but on accuracy its effect is small: disabling it matched or slightly improved validation accuracy in our runs. We keep it for geometric coverage; consistent with this, the interior-shell vertex positions move very little during training.

\subsection{Where the error is}
\label{sec:exp-error}

The per-class accuracy spread is class-structured and stable across runs:
vehicle and global-shape classes (automobile, truck, ship, frog) are classified
most reliably, while fine-texture animal classes (cat, bird, dog, deer) are
weakest, with cat the hardest. 

To locate the source of the gap we evaluate a separately converged checkpoint
(test accuracy $86.5\%$) on the training set with augmentation disabled. It
reaches $95.7\%$ on the clean training data, a gap of roughly nine points. The
model therefore fits most of clean CIFAR-10 through the spherical tokenizer,
though it does not interpolate (comparable ResNets reach nearly $100\%$ train
accuracy); the test gap appears dominated by generalization, with a smaller
fitting shortfall. In the same direction, preliminary runs varying the graph size and adding
mixup/cutmix regularization did not close the gap. The flat-lattice control (\S\ref{sec:exp-grid}) points the same way: in the matched attention model, replacing the spherical layout while keeping the rest of that pipeline fixed closes most of its gap to convolutional baselines. We discuss the implication in
\S\ref{sec:limitations}.

\section{Limitations}
\label{sec:limitations}

QSV does not match strong convolutional or transformer baselines trained from
scratch on CIFAR-10, which commonly exceed $90\%$; the flat-lattice control of \S\ref{sec:exp-grid}, run on the matched attention model, suggests much of this deficit traces to the spherical layout itself.
Our diagnostic in \S\ref{sec:exp-error} is consistent with this gap being
dominated by generalization: the model fits clean training data to
$95.7\%$ through the spherical tokenizer but tests at $86.5\%$. The errors are class-structured, with fine-texture animal classes weakest (\S\ref{sec:exp-error}), suggesting that
the spherical tokenizer and sparse graph transport preserve global object
structure better than fine local texture. On planar images the flat-lattice
variant is the natural remedy (\S\ref{sec:exp-grid}).

Two further caveats limit scope. First, our findings concern two
small natural-image datasets (CIFAR-10 and CIFAR-100) at one model scale; the
transport-versus-routing asymmetry replicates across both (with single runs per
CIFAR-100 variant), while the matched-attention comparison used three seeds on CIFAR-10 plus a
single-run CIFAR-100 baseline and value-path ablation, the flat-lattice
comparison used three seeds on CIFAR-10 only, and the joint-disable, frozen-field, and same-pipeline ResNet-20 cells are
single runs at one seed. We have not evaluated whether these findings persist at larger
scale, at higher resolution, on natively spherical data, or against stronger
baselines. Second, although the kernel is built from rotations, we do not claim that
the network is equivariant to input rotations: the base quaternions are free
parameters attached to fixed vertex positions and the perturbation MLP is not
equivariant, so the construction is better described as a learned quaternion
connection between per-vertex frames than as a rotation-equivariant operator.
Establishing what symmetry, if any, the learned connection preserves is left to
future work.

\section{Conclusion}
\label{conc}

We studied Quat-Sphere-Vision, a sparse spherical vision model in which a single learned unit quaternion per token acts, through one relative rotation, as both routing weight and sandwich-product feature transport.

The model trains reliably ($85.9\%$ default-configuration mean, $86.9\%$ best across the sweep on CIFAR-10), and its two kernel roles are unequal on both datasets (transport
$-4.3$/$-4.1$ points; routing within noise) yet jointly indispensable (both
disabled: $64.1$). Under matched controls, standard attention on the same graph exceeds QSV ($87.3\%$), and the flat lattice reaches $91.1\%$ against a same-pipeline ResNet-20's $93.2\%$ (single run). When one quaternion must serve both roles, the learned pairwise computation concentrates almost entirely in transport, giving up the redundancy that makes standard attention robust.

\paragraph{Future work.} Extending the matched comparison to equivariant-attention baselines~\cite{fuchs2020se3,brehmer2023gatr} and to the closest quaternion-routing prior~\cite{zhao2020qecnet}, and testing the asymmetry at larger scale.

\bibliography{references}

\begin{thebibliography}{47}
\providecommand{\natexlab}[1]{#1}

\bibitem[{B{\'e}cigneul and Ganea(2019)}]{becigneul2019rieadam}
B{\'e}cigneul, G.; and Ganea, O.-E. 2019.
\newblock Riemannian Adaptive Optimization Methods.
\newblock In \emph{International Conference on Learning Representations
  (ICLR)}.

\bibitem[{Bonnabel(2013)}]{bonnabel2013riemannian}
Bonnabel, S. 2013.
\newblock Stochastic Gradient Descent on {R}iemannian Manifolds.
\newblock \emph{IEEE Transactions on Automatic Control}, 58(9): 2217--2229.

\bibitem[{Brehmer et~al.(2023)Brehmer, de~Haan, Behrends, and
  Cohen}]{brehmer2023gatr}
Brehmer, J.; de~Haan, P.; Behrends, S.; and Cohen, T.~S. 2023.
\newblock Geometric Algebra Transformer.
\newblock In \emph{Advances in Neural Information Processing Systems
  (NeurIPS)}.

\bibitem[{Brody, Alon, and Yahav(2022)}]{brody2022gatv2}
Brody, S.; Alon, U.; and Yahav, E. 2022.
\newblock How Attentive are Graph Attention Networks?
\newblock In \emph{International Conference on Learning Representations
  (ICLR)}.

\bibitem[{Byerly, Kalganova, and Dear(2021)}]{byerly2021norouting}
Byerly, A.; Kalganova, T.; and Dear, I. 2021.
\newblock No Routing Needed Between Capsules.
\newblock \emph{Neurocomputing}, 463: 545--553.

\bibitem[{Carlsson et~al.(2024)Carlsson, Gerken, Linander, Spie{\ss}, Ohlsson,
  Petersson, and Persson}]{carlsson2024healswin}
Carlsson, O.; Gerken, J.~E.; Linander, H.; Spie{\ss}, H.; Ohlsson, F.;
  Petersson, C.; and Persson, D. 2024.
\newblock {HEAL-SWIN}: A Vision Transformer on the Sphere.
\newblock In \emph{Proceedings of the IEEE/CVF Conference on Computer Vision
  and Pattern Recognition (CVPR)}.

\bibitem[{Choe et~al.(2022)Choe, Park, Rameau, Park, and
  Kweon}]{choe2022pointmixer}
Choe, J.; Park, C.; Rameau, F.; Park, J.; and Kweon, I.~S. 2022.
\newblock PointMixer: {MLP}-Mixer for Point Cloud Understanding.
\newblock In \emph{European Conference on Computer Vision (ECCV)}.

\bibitem[{Cohen et~al.(2018)Cohen, Geiger, K{\"o}hler, and
  Welling}]{cohen2018sphericalcnn}
Cohen, T.~S.; Geiger, M.; K{\"o}hler, J.; and Welling, M. 2018.
\newblock Spherical {CNN}s.
\newblock In \emph{International Conference on Learning Representations
  (ICLR)}.

\bibitem[{Cohen et~al.(2019)Cohen, Weiler, Kicanaoglu, and
  Welling}]{cohen2019gauge}
Cohen, T.~S.; Weiler, M.; Kicanaoglu, B.; and Welling, M. 2019.
\newblock Gauge Equivariant Convolutional Networks and the Icosahedral {CNN}.
\newblock In \emph{International Conference on Machine Learning (ICML)}.

\bibitem[{de~Haan et~al.(2021)de~Haan, Weiler, Cohen, and
  Welling}]{dehaan2021gem}
de~Haan, P.; Weiler, M.; Cohen, T.; and Welling, M. 2021.
\newblock Gauge Equivariant Mesh {CNNs}: Anisotropic Convolutions on Geometric
  Graphs.
\newblock In \emph{International Conference on Learning Representations
  (ICLR)}.

\bibitem[{Dong, Cordonnier, and Loukas(2021)}]{dong2021rank}
Dong, Y.; Cordonnier, J.-B.; and Loukas, A. 2021.
\newblock Attention is Not All You Need: Pure Attention Loses Rank Doubly
  Exponentially with Depth.
\newblock In \emph{International Conference on Machine Learning (ICML)}.

\bibitem[{Dong et~al.(2025)Dong, Noci, Khodak, and Li}]{dong2025mixit}
Dong, Y.; Noci, L.; Khodak, M.; and Li, M. 2025.
\newblock Is Random Attention Sufficient for Sequence Modeling? Disentangling
  Trainable Components in the Transformer.
\newblock \emph{arXiv preprint arXiv:2506.01115}.

\bibitem[{Dosovitskiy et~al.(2021)Dosovitskiy, Beyer, Kolesnikov, Weissenborn,
  Zhai, Unterthiner, Dehghani, Minderer, Heigold, Gelly, Uszkoreit, and
  Houlsby}]{dosovitskiy2021vit}
Dosovitskiy, A.; Beyer, L.; Kolesnikov, A.; Weissenborn, D.; Zhai, X.;
  Unterthiner, T.; Dehghani, M.; Minderer, M.; Heigold, G.; Gelly, S.;
  Uszkoreit, J.; and Houlsby, N. 2021.
\newblock An Image is Worth 16x16 Words: Transformers for Image Recognition at
  Scale.
\newblock In \emph{International Conference on Learning Representations
  (ICLR)}.

\bibitem[{Fuchs et~al.(2020)Fuchs, Worrall, Fischer, and
  Welling}]{fuchs2020se3}
Fuchs, F.~B.; Worrall, D.~E.; Fischer, V.; and Welling, M. 2020.
\newblock {SE(3)}-Transformers: {3D} Roto-Translation Equivariant Attention
  Networks.
\newblock In \emph{Advances in Neural Information Processing Systems
  (NeurIPS)}.

\bibitem[{Garc{\'i}a-Castellanos, Weiler, and
  Bekkers(2026)}]{garciacastellanos2026rove}
Garc{\'i}a-Castellanos, A.; Weiler, M.; and Bekkers, E.~J. 2026.
\newblock {RoVE}: Rotary Value Embeddings Attention for Relative
  Position-dependent Value Pathways.
\newblock \emph{arXiv preprint arXiv:2606.11275}.

\bibitem[{Gu and Tresp(2020)}]{gu2020capsules}
Gu, J.; and Tresp, V. 2020.
\newblock Improving the Robustness of Capsule Networks to Image Affine
  Transformations.
\newblock In \emph{Proceedings of the IEEE/CVF Conference on Computer Vision
  and Pattern Recognition (CVPR)}.

\bibitem[{Hassani et~al.(2021)Hassani, Walton, Shah, Abuduweili, Li, and
  Shi}]{hassani2021cct}
Hassani, A.; Walton, S.; Shah, N.; Abuduweili, A.; Li, J.; and Shi, H. 2021.
\newblock Escaping the Big Data Paradigm with Compact Transformers.
\newblock \emph{arXiv preprint arXiv:2104.05704}.

\bibitem[{He et~al.(2016)He, Zhang, Ren, and Sun}]{he2016resnet}
He, K.; Zhang, X.; Ren, S.; and Sun, J. 2016.
\newblock Deep Residual Learning for Image Recognition.
\newblock In \emph{Proceedings of the IEEE Conference on Computer Vision and
  Pattern Recognition (CVPR)}, 770--778.

\bibitem[{Heo et~al.(2024)Heo, Park, Han, and Yun}]{heo2024ropevit}
Heo, B.; Park, S.; Han, D.; and Yun, S. 2024.
\newblock Rotary Position Embedding for Vision Transformer.
\newblock In \emph{European Conference on Computer Vision (ECCV)}.

\bibitem[{Hyeon-Woo et~al.(2023)Hyeon-Woo, Yu-Ji, Heo, Han, Oh, and
  Oh}]{hyeonwoo2023scratching}
Hyeon-Woo, N.; Yu-Ji, K.; Heo, B.; Han, D.; Oh, S.~J.; and Oh, T.-H. 2023.
\newblock Scratching Visual Transformer's Back with Uniform Attention.
\newblock In \emph{Proceedings of the IEEE/CVF International Conference on
  Computer Vision (ICCV)}.

\bibitem[{Ji(2025)}]{ji2025riemannformer}
Ji, Z. 2025.
\newblock {RiemannFormer}: A Framework for Attention in Curved Spaces.
\newblock \emph{arXiv preprint arXiv:2506.07405}.

\bibitem[{Ji(2026)}]{ji2026cliffordnet}
Ji, Z. 2026.
\newblock {CliffordNet}: All You Need is Geometric Algebra.
\newblock \emph{arXiv preprint arXiv:2601.06793}.

\bibitem[{Krizhevsky(2009)}]{krizhevsky2009cifar}
Krizhevsky, A. 2009.
\newblock Learning Multiple Layers of Features from Tiny Images.
\newblock Technical report, University of Toronto.

\bibitem[{Liao and Smidt(2023)}]{liao2023equiformer}
Liao, Y.-L.; and Smidt, T. 2023.
\newblock Equiformer: Equivariant Graph Attention Transformer for {3D}
  Atomistic Graphs.
\newblock In \emph{International Conference on Learning Representations
  (ICLR)}.

\bibitem[{Loshchilov and Hutter(2019)}]{loshchilov2019adamw}
Loshchilov, I.; and Hutter, F. 2019.
\newblock Decoupled Weight Decay Regularization.
\newblock In \emph{International Conference on Learning Representations
  (ICLR)}.

\bibitem[{M{\"u}ller and Hutter(2021)}]{muller2021trivialaugment}
M{\"u}ller, S.~G.; and Hutter, F. 2021.
\newblock {TrivialAugment}: Tuning-Free Yet State-of-the-Art Data Augmentation.
\newblock In \emph{Proceedings of the IEEE/CVF International Conference on
  Computer Vision (ICCV)}.

\bibitem[{Nguyen, Nguyen, and Phung(2021)}]{nguyen2021qgnn}
Nguyen, D.~Q.; Nguyen, T.~D.; and Phung, D. 2021.
\newblock Quaternion Graph Neural Networks.
\newblock In \emph{Asian Conference on Machine Learning (ACML)}, Proceedings of
  Machine Learning Research.

\bibitem[{Paik, Kwak, and Kim(2019)}]{paik2019capsules}
Paik, I.; Kwak, T.; and Kim, I. 2019.
\newblock Capsule Networks Need an Improved Routing Algorithm.
\newblock In \emph{Asian Conference on Machine Learning (ACML)}, Proceedings of
  Machine Learning Research.

\bibitem[{Parcollet, Morchid, and Linar{\`e}s(2020)}]{parcollet2020survey}
Parcollet, T.; Morchid, M.; and Linar{\`e}s, G. 2020.
\newblock A Survey of Quaternion Neural Networks.
\newblock \emph{Artificial Intelligence Review}, 53: 2957--2982.

\bibitem[{Parcollet et~al.(2018)Parcollet, Zhang, Morchid, Trabelsi,
  Linar{\`e}s, De~Mori, and Bengio}]{parcollet2018qcnn}
Parcollet, T.; Zhang, Y.; Morchid, M.; Trabelsi, C.; Linar{\`e}s, G.; De~Mori,
  R.; and Bengio, Y. 2018.
\newblock Quaternion Convolutional Neural Networks for End-to-End Automatic
  Speech Recognition.
\newblock In \emph{Interspeech}.

\bibitem[{Raghu et~al.(2021)Raghu, Unterthiner, Kornblith, Zhang, and
  Dosovitskiy}]{raghu2021vision}
Raghu, M.; Unterthiner, T.; Kornblith, S.; Zhang, C.; and Dosovitskiy, A. 2021.
\newblock Do Vision Transformers See Like Convolutional Neural Networks?
\newblock In \emph{Advances in Neural Information Processing Systems
  (NeurIPS)}, 12116--12128.

\bibitem[{Sabour, Frosst, and Hinton(2017)}]{sabour2017capsules}
Sabour, S.; Frosst, N.; and Hinton, G.~E. 2017.
\newblock Dynamic Routing Between Capsules.
\newblock In \emph{Advances in Neural Information Processing Systems
  (NeurIPS)}.

\bibitem[{Saff and Kuijlaars(1997)}]{saff1997distributing}
Saff, E.~B.; and Kuijlaars, A. B.~J. 1997.
\newblock Distributing Many Points on a Sphere.
\newblock \emph{The Mathematical Intelligencer}, 19(1): 5--11.

\bibitem[{Shen et~al.(2020)Shen, Zhang, Huang, Wei, and
  Zhang}]{shen2020rotationequiv}
Shen, W.; Zhang, B.; Huang, S.; Wei, Z.; and Zhang, Q. 2020.
\newblock {3D}-Rotation-Equivariant Quaternion Neural Networks.
\newblock In \emph{European Conference on Computer Vision (ECCV)}.

\bibitem[{Su et~al.(2024)Su, Ahmed, Lu, Pan, Bo, and Liu}]{su2024rope}
Su, J.; Ahmed, M.; Lu, Y.; Pan, S.; Bo, W.; and Liu, Y. 2024.
\newblock {RoFormer}: Enhanced Transformer with Rotary Position Embedding.
\newblock \emph{Neurocomputing}, 568: 127063.

\bibitem[{Tay et~al.(2021)Tay, Bahri, Metzler, Juan, Zhao, and
  Zheng}]{tay2021synthesizer}
Tay, Y.; Bahri, D.; Metzler, D.; Juan, D.-C.; Zhao, Z.; and Zheng, C. 2021.
\newblock Synthesizer: Rethinking Self-Attention for Transformer Models.
\newblock In \emph{International Conference on Machine Learning (ICML)}.

\bibitem[{Tay et~al.(2019)Tay, Zhang, Luu, Rao, Zhang, Wang, Fu, and
  Hui}]{tay2019quaternion}
Tay, Y.; Zhang, A.; Luu, A.~T.; Rao, J.; Zhang, S.; Wang, S.; Fu, J.; and Hui,
  S.~C. 2019.
\newblock Lightweight and Efficient Neural Natural Language Processing with
  Quaternion Networks.
\newblock In \emph{Annual Meeting of the Association for Computational
  Linguistics (ACL)}.

\bibitem[{Tolstikhin et~al.(2021)Tolstikhin, Houlsby, Kolesnikov, Beyer, Zhai,
  Unterthiner, Yung, Steiner, Keysers, Uszkoreit, Lucic, and
  Dosovitskiy}]{tolstikhin2021mlpmixer}
Tolstikhin, I.; Houlsby, N.; Kolesnikov, A.; Beyer, L.; Zhai, X.; Unterthiner,
  T.; Yung, J.; Steiner, A.; Keysers, D.; Uszkoreit, J.; Lucic, M.; and
  Dosovitskiy, A. 2021.
\newblock {MLP}-Mixer: An all-{MLP} Architecture for Vision.
\newblock In \emph{Advances in Neural Information Processing Systems
  (NeurIPS)}.

\bibitem[{Vaswani et~al.(2017)Vaswani, Shazeer, Parmar, Uszkoreit, Jones,
  Gomez, Kaiser, and Polosukhin}]{vaswani2017attention}
Vaswani, A.; Shazeer, N.; Parmar, N.; Uszkoreit, J.; Jones, L.; Gomez, A.~N.;
  Kaiser, {\L}.; and Polosukhin, I. 2017.
\newblock Attention Is All You Need.
\newblock In \emph{Advances in Neural Information Processing Systems
  (NeurIPS)}.

\bibitem[{Yamauchi, Nitta, and Tamori(2026)}]{yamauchi2026quaternion}
Yamauchi, S.; Nitta, T.; and Tamori, H. 2026.
\newblock Quaternion Self-Attention with Shared Scores.
\newblock In \emph{International Conference on Machine Learning (ICML)}.
\newblock ArXiv:2605.24920.

\bibitem[{Yao and Yang(2025)}]{yao2025geope}
Yao, Y.; and Yang, B. 2025.
\newblock {GeoPE}: A Unified Geometric Positional Embedding for Structured
  Tensors.
\newblock \emph{arXiv preprint arXiv:2512.04963}.

\bibitem[{Yu et~al.(2022)Yu, Luo, Zhou, Si, Zhou, Wang, Feng, and
  Yan}]{yu2022metaformer}
Yu, W.; Luo, M.; Zhou, P.; Si, C.; Zhou, Y.; Wang, X.; Feng, J.; and Yan, S.
  2022.
\newblock MetaFormer Is Actually What You Need for Vision.
\newblock In \emph{Proceedings of the IEEE/CVF Conference on Computer Vision
  and Pattern Recognition (CVPR)}, 10819--10829.

\bibitem[{Zhai et~al.(2023)Zhai, Likhomanenko, Littwin, Busbridge, Ramapuram,
  Zhang, Gu, and Susskind}]{zhai2023stabilizing}
Zhai, S.; Likhomanenko, T.; Littwin, E.; Busbridge, D.; Ramapuram, J.; Zhang,
  Y.; Gu, J.; and Susskind, J.~M. 2023.
\newblock Stabilizing Transformer Training by Preventing Attention Entropy
  Collapse.
\newblock In \emph{International Conference on Machine Learning (ICML)}, volume
  202 of \emph{Proceedings of Machine Learning Research}, 40770--40803.

\bibitem[{Zhang et~al.(2021)Zhang, Tay, Zhang, Chan, Luu, Hui, and
  Fu}]{zhang2021phm}
Zhang, A.; Tay, Y.; Zhang, S.; Chan, A.; Luu, A.~T.; Hui, S.~C.; and Fu, J.
  2021.
\newblock Beyond Fully-Connected Layers with Quaternions: Parameterization of
  Hypercomplex Multiplications with $1/n$ Parameters.
\newblock In \emph{International Conference on Learning Representations
  (ICLR)}.

\bibitem[{Zhang et~al.(2019)Zhang, Tay, Yao, and Liu}]{zhang2019quate}
Zhang, S.; Tay, Y.; Yao, L.; and Liu, Q. 2019.
\newblock Quaternion Knowledge Graph Embeddings.
\newblock In \emph{Advances in Neural Information Processing Systems
  (NeurIPS)}, 2731--2741.

\bibitem[{Zhao et~al.(2021)Zhao, Jiang, Jia, Torr, and
  Koltun}]{zhao2021pointtransformer}
Zhao, H.; Jiang, L.; Jia, J.; Torr, P. H.~S.; and Koltun, V. 2021.
\newblock Point Transformer.
\newblock In \emph{IEEE/CVF International Conference on Computer Vision
  (ICCV)}.

\bibitem[{Zhao et~al.(2020)Zhao, Birdal, Lenssen, Menegatti, Guibas, and
  Tombari}]{zhao2020qecnet}
Zhao, Y.; Birdal, T.; Lenssen, J.~E.; Menegatti, E.; Guibas, L.; and Tombari,
  F. 2020.
\newblock Quaternion Equivariant Capsule Networks for {3D} Point Clouds.
\newblock In \emph{European Conference on Computer Vision (ECCV)}.

\end{thebibliography}

\appendix
\section{Default Configuration}
\label{sec:appendix-config}

The default values used in our CIFAR-10 experiments are: feature dimension $d = 256$; number of heads $H = 16$ (giving $Q = d/(H\cdot 4) = 4$ quaternion channels per head); base shell vertex count $V_1 = 128$; number of shells $L = 5$; shell vertex counts $V = [128, 192, 256, 320, 384]$; per shell neighbor counts $K = [14, 14, 16, 16, 16]$; resulting edge ratio $\approx 0.05$ (per-shell dense denominator); batch size $128$; up to $500$ epochs with early stopping on validation accuracy (patience $50$ epochs, minimum improvement $10^{-4}$). Test accuracy is reported at the best-validation checkpoint.

\paragraph{Optimization.} AdamW (weight decay $0.05$) with a one-cycle cosine-shaped schedule: peak learning rate $10^{-3}$, $10\%$ warmup, initial division factor $10$, final division factor $100$; position learning rate $5\times10^{-5}$ (no weight decay); the $\hat\varepsilon$ logits train at $3\times$ the base rate. The quaternion group uses Riemannian Adam at a constant $2.5\times10^{-3}$.

\paragraph{Regularization.} Label smoothing $0.1$; attention and output dropout $0.1$; quaternion smoothness coefficient $\xi = 10^{-3}$; head-diversity coefficient $\lambda_{\mathrm{div}} = 10^{-2}$; $\ell_2$ penalty on the $\hat\beta$ logits $10^{-5}$. Augmentation: $4$-pixel-padded random crop, horizontal flip, TrivialAugmentWide.

\paragraph{Geometry.} The perturbation bound $\varepsilon_l$ is sigmoid-parameterized in $(0.01, \pi/4)$ per shell. Sphere repulsion uses $k=8$ neighbors ($16$ on CIFAR-100), applied every $10$ batches with a target step decayed to zero over roughly the first $40\%$ of training. Training is in single precision (fp32), with \texttt{torch.compile} enabled.

\paragraph{Per-seed values (Table~\ref{tab:ablation}).} CIFAR-10 per-seed
test accuracies: baseline $85.9/86.0$; transport~$\to$~identity
$81.3/81.7/82.1$; uniform $85.3/86.1$; $|\Re(r)|$ $86.0/86.5$; geometric
$85.4/86.5$; attention $W_V \to$ identity $86.2/86.3/86.5$ (mean over three
seeds).

\paragraph{Compute and code.} On the same RTX A6000 (fp32, identical data
pipeline), compiled QSV trains at ${\approx}470$ images/s (a full run is about
$14$ hours); the matched attention baseline reaches ${\approx}1550$ images/s
uncompiled and ${\approx}5350$ images/s compiled. Code and configurations will
be released upon publication.

\end{document}